\documentclass[conference]{IEEEconf}
\IEEEoverridecommandlockouts

\usepackage{cite}
\usepackage{amsmath,amssymb,amsfonts}
\usepackage[ruled,vlined]{algorithm2e}
\usepackage{graphicx}
\usepackage{textcomp}
\usepackage{xcolor}
\usepackage{booktabs}
\usepackage[colorlinks]{hyperref}
\usepackage{pifont}
\usepackage[normalem]{ulem}
\newcommand{\cmark}{\ding{51}}
\newcommand{\xmark}{\ding{55}}
\def\BibTeX{{\rm B\kern-.05em{\sc i\kern-.025em b}\kern-.08em
    T\kern-.1667em\lower.7ex\hbox{E}\kern-.125emX}}
\begin{document}

\newcommand{\blue}[1]{
{\color{blue}{#1}}
}

\newcommand{\red}[1]{
{\color{red}{#1}}
}

\newcommand{\green}[1]{
{\color{green}{#1}}
}


\title{\LARGE \bf Real-Time Gait Adaptation for Quadrupeds using Model Predictive Control and  Reinforcement Learning






}

\author{
Prakrut Kotecha$^*$$^{1}$, Ganga Nair B$^*$$^{1}$, Shishir Kolathaya$^{2}$
\thanks{$^*$ Equal Contribution. Decided by cycle race}
\thanks{$^{1}$G. Nair, P. Kotecha, are with the Robert Bosch Center for Cyber-Physical Systems, Indian Institute of Science, Bengaluru.}%
\thanks{$^{2}$S. Kolathaya is with the Robert Bosch Center for Cyber-Physical Systems and the Department of Computer Science \& Automation, Indian Institute of Science, Bengaluru.}
\thanks{This project is funded by ARTPARK}
\thanks{Email: \href{mailto:stochlab@iisc.ac.in}{stochlab@iisc.ac.in}, Website: \href{https://sites.google.com/iisc.ac.in/autogait/home}{link}}%
}
\maketitle

\begin{abstract}

Model-free reinforcement learning (RL) has enabled adaptable and agile quadruped locomotion; however, policies often converge to a single gait, leading to suboptimal performance. Traditionally, Model Predictive Control (MPC) has been extensively used to obtain task-specific optimal policies but lacks the ability to adapt to varying environments. To address these limitations, we propose an optimization framework for real-time gait adaptation in a continuous gait space, combining the Model Predictive Path Integral (MPPI) algorithm with a Dreamer module to produce adaptive and optimal policies for quadruped locomotion. At each time step, MPPI jointly optimizes the actions and gait variables using a learned Dreamer reward that promotes velocity tracking, energy efficiency, stability, and smooth transitions, while penalizing abrupt gait changes. A learned value function is incorporated as terminal reward, extending the formulation to an infinite-horizon planner. We evaluate our framework in simulation on the Unitree Go1, demonstrating an average reduction of up to 36.48\% in energy consumption across varying target speeds, while maintaining accurate tracking and adaptive, task-appropriate gaits.

\end{abstract}
\vspace{-10pt}
\section{Introduction}

In nature, animals instinctively transition between gaits—such as walking, trotting, and galloping—based on environmental context and task demands. This behavior is often attributed to principles of energy minimization~\cite{hoyt1981gait}, dynamic stability~\cite{granatosky2018stride}, and robustness to perturbations~\cite{shafiee2024viability}.
Quadrupedal robots have achieved remarkable progress in replicating the agility and versatility of animals, including dynamic motions like jumping or parkour~\cite{Hoeller2023ANYmalPL}.
However, most research efforts for multi-terrain locomotion have resulted in singular gait policies—typically trotting or walking, or in diverse but preprogrammed or commanded behaviors.
Therefore, developing strategies for adaptive gait selection and seamless transitions between gaits represents a natural and essential next step in advancing quadrupedal locomotion.



\begin{figure}
    \centering
    \includegraphics[width=0.9\linewidth]{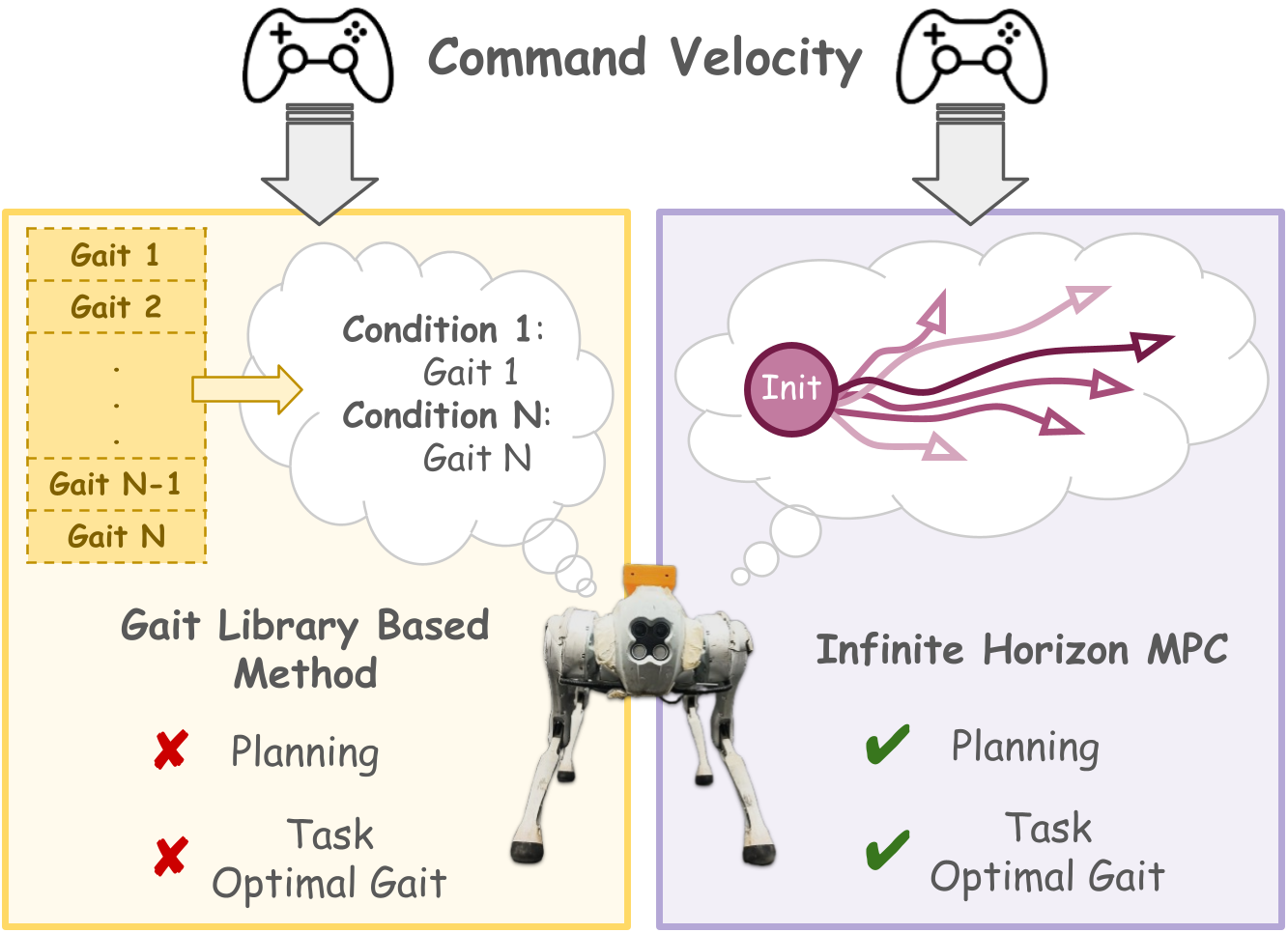}
    \caption{Conventional approaches (left) rely on discrete gait switching with predefined libraries. Our framework (right) performs joint optimization over actions and continuous gait parameters using learned dynamics and a reward-based planner. \vspace{-10pt}}
    \label{fig:concept image}
\end{figure}

Traditionally, Model Predictive Control (MPC) has been used to generate complex, dynamically consistent motion~\cite{Carlo2018DynamicLI,Raiola2020ASY,Kim2019HighlyDQ}. These approaches typically assume fixed gait parameters—such as contact sequences or timing offsets—defined \textit{a priori}. This assumption limits flexibility during deployment, particularly under changing task demands or varying terrains. Moreover, MPC requires accurate dynamics models and can be computationally demanding, making it less suited for online adaptation in complex environments.

To overcome these modeling and adaptability limitations, Reinforcement Learning (RL) provides a data-driven alternative that learns control policies directly from interaction, without requiring an explicit model of the robot~\cite{eth2019}. While RL enables emergent locomotion behaviors, designing reward functions that yield diverse and stable gaits remains non-trivial. Simple rewards often produce unstable or repetitive behaviors, whereas complex ones tend to bias the policy toward specific gaits.

Building on this, \textit{Hierarchical RL} methods~\cite{FastAE,Wei2023LearningMQ} have been proposed to handle multi-gait behaviors more explicitly. These architectures decompose control into a high-level gait selector and a low-level motion controller, improving behavioral structure and interpretability. However, such frameworks often interpolate between similar symmetric gaits (e.g., walk$\leftrightarrow$trot) and struggle to generalize beyond predefined patterns.



These limitations motivate hybrid strategies that integrate the structure and stability of model-based control with the adaptability of model-free learning. In particular, combining RL with MPC has shown improved performance in complex dynamical systems~\cite{piploco,loop,tdmpc2}. During training, such methods commonly employ a Dreamer-like architecture~\cite{dreamer} to jointly learn a dynamics model, reward estimator, policy, and value function for long-horizon control. During inference, an infinite-horizon trajectory optimization problem is solved by warm-starting the process using the RL policy, enabling faster convergence and greater adaptability. However, these approaches still inherit a key limitation of RL-based policies—namely, the lack of gait , as the underlying RL policy typically converges to a single dominant behavior.


\begin{table}[htb!]
\centering
\caption{Comparison of gait selection methods.}
\label{tab:method_comparison}
\scriptsize
\setlength{\tabcolsep}{3pt}
\renewcommand{\arraystretch}{1.3}
\begin{tabular}{l|c|c|c|c|c}
\toprule
\textbf{Capability} & \textbf{End-to-End} & \textbf{Classical} & \textbf{Gait} & \textbf{Dreamer} & \textbf{Ours} \\
 & \textbf{RL} & \textbf{MPC} & \textbf{Library} & \textbf{Based} & \textbf{} \\
 & \textbf{\cite{eth2019}} & \textbf{\cite{Carlo2018DynamicLI}} & \textbf{\cite{Amatucci2022MonteCT}} & \textbf{\cite{piploco}} & \textbf{} \\
\midrule
Gait Diversity & \cmark & \xmark & \cmark & \xmark & \cmark \\
Smooth Transitions & \cmark & \xmark & \xmark & \xmark & \cmark \\
Online Gait Adaptation & \xmark & \xmark & \cmark & \xmark & \cmark \\
Dynamics Awareness & \xmark & \cmark & \xmark & \cmark & \cmark \\
Planning & \xmark & \cmark & \xmark & \cmark & \cmark \\
\bottomrule
\end{tabular}
\vspace{-10pt}
\end{table}

As evident from Table~\ref{tab:method_comparison}, none of the existing approaches provide all the desired capabilities within a single, unified framework. To address these challenges, we propose a control framework for quadrupeds that enables \textit{continuous, real-time gait adaptation} by jointly optimizing both actions and gait parameters. Inspired by recent work~\cite{piploco}, our method integrates Model Predictive Path Integral (MPPI) control with a Dreamer module trained on diverse gait data, which supplies learned dynamics, reward, value, and policy priors to guide the optimization.

\textbf{Key Contributions}
\begin{itemize}
    \item We introduce a novel control framework that performs \textit{infinite-horizon planning over gait and actions} via MPPI guided by a learned dreamer module, enabling diverse gaits, smooth transitions and adaptive behavior in real time.
    \item Our method achieves an average reduction of up to \textbf{36.48 \%} in energy consumption compared to baseline approaches across varying target speeds, while maintaining accurate tracking and robust gait switching.
    \item The framework is modular and can be used with any RL algorithm supporting gait-conditioned policies.
    
\end{itemize}
\section{Setup}  
\label{sc:setup}

\vspace{-5pt}
In this section, we describe the approach incorporated, starting with the problem formulation and then detailing the training and deployment processes. To support planning under unknown dynamics, we utilize a dreamer module trained in parallel with the RL policy. This module learns all the parameters required by the MPPI planner during deployment. While the RL policy is trained using gait-specific rewards to induce diverse locomotion behaviors, the dreamer reward is learned using task and energy-specific terms, the details of which are explained in section \ref{rew}.

Our implementation builds on the Walk These Ways framework~\cite{wtw}, a gait-dependent RL architecture in which the policy is conditioned on manually specified gait parameters. While we adopt this structure in our work, the proposed planning framework is modular and readily extensible to other gait-dependent locomotion architectures.

\subsection{Preliminaries}  

Quadruped locomotion is modeled as an infinite-horizon discounted Partially Observable Markov Decision Process (POMDP), defined by the tuple:  
\begin{equation}  
\mathcal{M} = \{\mathcal{S}, \mathcal{A}, \mathcal{O}, r, \mathcal{P}, \gamma, \mathcal{F}\}
\end{equation}  
where $\mathcal{S} \subset \mathbb{R}^n$ represents the state space, $\mathcal{O} \subset \mathbb{R}^p$ the observation space, and $\mathcal{A} \subset \mathbb{R}^m$ the action space. The system dynamics are governed by the transition function $\mathcal{P}: \mathcal{S} \times \mathcal{A} \mapsto \Pr(\cdot)$, with $\Pr$ being a probability measure, while the observation function $\mathcal{F}: \mathcal{S} \mapsto \mathcal{O}$ provides partial observability. The reward function $r: \mathcal{S} \times \mathcal{A} \mapsto \mathbb{R}$ guides the learning process, and $\gamma \in (0,1)$ is the discount factor.

\subsection{Training} \label{Training}

\begin{figure}
    \centering    
    \includegraphics[width=0.9\linewidth]{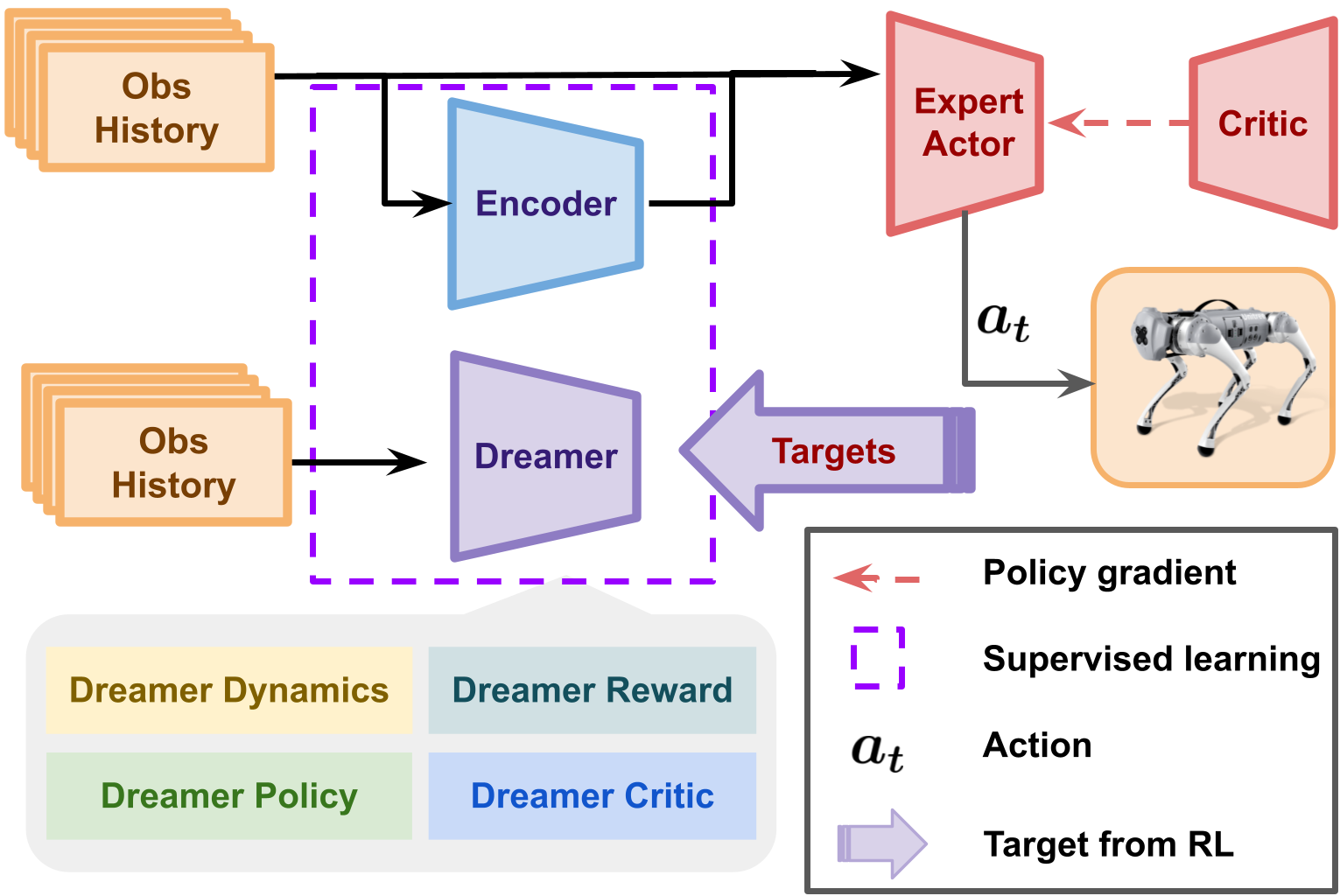}
    \caption{Overview of the training pipeline: the dreamer module learns latent dynamics, reward, and value functions from observation histories, while the expert actor is trained using supervised signals from dreamer outputs. \vspace{-10pt}}
    \label{fig:training}
\end{figure}

To solve this POMDP in a RL context, we employ an Actor-Critic framework trained using the Proximal Policy Optimization (PPO) algorithm \cite{ppo, wtw}. In parallel, the adaptation module $\mathcal{E_\theta}$ (encoder) and the dreamer module are trained using supervised learning (see fig. \ref{fig:training}). The adaptation module is trained to estimate privileged information from observation history, which is then used to assist the actor during deployment \cite{wtw}.





\textbf{Observation \& Action Space:} The observations in our task includes the robot’s joint positions relative to their nominal values, joint velocities, the timing reference variables and commanded gait ($g_{cmd}$), projected gravity, and previous two actions. The action space consists of joint-angle perturbations ($a_t$) applied to the nominal joint angles ($q_{\text{nominal}}$), where the commanded joint position ($q_t$) at time $t$ is given by: $q_t = a_t + q_{\text{nominal}}$. These commands are processed through an actuator network \cite{eth2019} to compute the torques required for execution.  

\textbf{Reward:} For training, we adopt the reward structure from Walk These Ways~\cite{wtw}, which combines task rewards with a range of auxiliary terms to enforce stable, gait-specific locomotion. These include penalties and incentives for tracking commanded velocities, maintaining posture, and achieving desired contact schedules.

\textbf{Dreamer}: Inspired by prior work in combining RL with planning~\cite{piploco}, we incorporate a dreamer module that learns all components necessary for deployment-time planning. This modular framework is trained in parallel with the RL policy and includes:

\begin{itemize}
\item \textbf{Policy} ($\pi_\theta$): A cloned policy trained to imitate the expert RL policy by minimizing the difference in actions under identical observations. This is used to warm-start the MPPI planner.  It takes a 30-step observation history ($o^{hist}_t$) as input and outputs action ($a_{cloned}$).

\item \textbf{Dynamics Model} ($D_\theta$): Learned from environment interactions to predict the next observation given the current observation and action.

\item \textbf{Reward Model} ($R_\theta$): A learned reward function trained to capture general performance metrics such as velocity tracking and energy efficiency. It takes observation and action as input.

\item \textbf{Value Function} ($V_\theta$): Estimates the expected cumulative reward under $R_\theta$, enabling infinite-horizon approximation without long rollouts during deployment. Unlike the RL critic, this is designed specifically for planning (see Section~\ref{rew}).

\end{itemize}





\section{Methodology}
\label{Method}

In this section, we present the details of our framework that enables gait optimization. The dreamer module, introduced in the previous section, is used by the Model Predictive Path Integral (MPPI) planner to optimize both control actions and gait parameters. We now describe the reward formulation used for learning, the structure of our MPPI-based planning algorithm, and the integrated framework as proposed above.

\subsection{Reward Formulation}
\label{rew}
During the training of RL policy for different gait commands, the critic may rely on gait-specific rewards (such as Raibert rewards \cite{raibert}) to enable effective learning of each gait’s locomotion strategy. However, to enable gait adaptation during deployment, we require a more generalized reward function—one that prioritizes energy efficiency and velocity tracking, without introducing bias toward any particular gait. To achieve this, we train a separate dreamer critic alongside the RL critic.

Table~\ref{tab:rewards} lists the components of the reward used during training and deployment. The first four terms are learned using dreamer and optimized jointly during training. The final two terms are added only during MPPI deployment to encourage smooth gait transitions and reduce actor divergence.

\begin{table}[htp!]
    \vspace{10pt}
    \centering
    \caption{Reward components used in training and deployment.}
    \label{tab:rewards}
    \scriptsize
    \setlength{\tabcolsep}{5pt}
    \renewcommand{\arraystretch}{1.3}
    \begin{tabular}{l|c|c}
        \toprule
        \textbf{Term} & \textbf{Description} & \textbf{Equation} \\
        \midrule
        $r_{vel}$ & Velocity tracking & 
        $-\left\|\mathbf{v}_{\text{robot}} - \mathbf{v}_{\text{target}}\right\|^2$  \\
        
        $r_{energy}$ & Energy efficiency & 
        $- \sum_t \| \boldsymbol{\tau}_t \cdot \dot{\mathbf{q}}_t \|$ \\
        
        $r_{ang}$ & Angular velocity & 
        $- \left\|\boldsymbol{\omega}_{\text{robot}}\right\|^2$  \\
        
        $r_{cont}$ & Action continuity & 
        $- \left\|\mathbf{a}_t - \mathbf{a}_{t-1}\right\|^2$  \\

        $r_{div}$ & Actor divergence & 
        $- \left\|\mathbf{a}_t - \mathbf{a}_{cloned}\right\|^2$ \\
        
        $r_{gait}$ & Gait continuity (Deployment only) & 
        $- \left\|g^{cmd}_{(t)} - g^{cmd}_{(t-1)}\right\|^2$ \\
        \bottomrule
    \end{tabular}
\end{table}


The total reward at time step ($t$) used during MPPI planning is given below. The weights for each rewards are $\alpha_1 - \alpha_6$ respectively (can be tuned according to requirements). 
\[
R_t = \alpha_1 r_{vel} + \alpha_2 r_{energy} + \alpha_3 r_{ang} + \alpha_4 r_{cont} + \alpha_5 r_{div} + \alpha_6 r_{gait}
\]

\begin{figure}
    \centering    
    \includegraphics[width=0.9\linewidth]{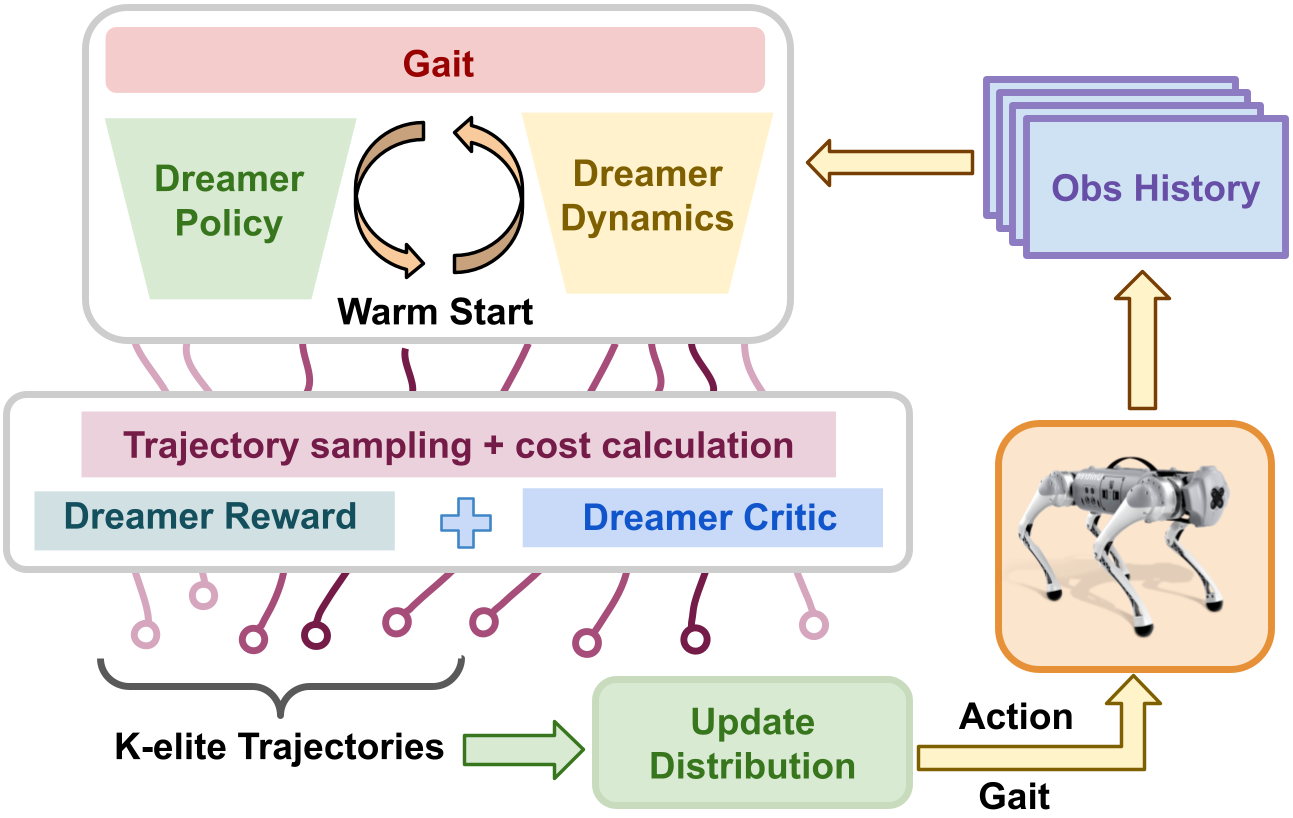}
    \caption{Pictorial representation of the MPPI algorithm: the warm start utilises dreamer policy and dynamics to generate initial trajectories to help MPPI.}
    \label{fig:deploy}
    \vspace{-10pt}
\end{figure}
\vspace{-10pt}
\subsection{Planning via MPPI}
We now shift our focus to the deployment phase (as shown in fig. \ref{fig:deploy}), where control actions and gait parameters are optimized in real time. Performing this optimization online enables the system to use up-to-date observations and to incorporate deployment-time constraints—such as physical limitations or task-specific objectives—directly into the decision-making process.

Our framework involves solving the following optimization problem to obtain the optimal action sequence $a_{t:t+H}^*$ and the gait commands $g^{*\text{cmd}}_{t:t+H}$ at the current observation ${o}_t$:

\vspace{-10pt}
\begin{equation}
\begin{aligned}
    \max_{\substack{(a_t, \dots, a_{t+H})\\(g^{\text{cmd}}_{t}, \dots, g^{\text{cmd}} _{t+H})}} \quad 
    & \sum_{k=0}^{H-1} \gamma^k R_{\theta}({\hat o}_{t+k}, a_{t+k}) 
    + \gamma^H V_{\theta}({\hat o}_{t+H}) \\
    \text{s.t.} \quad 
    & \hat o_{t} = o_t, \\
     {\hat o}_{t+k+1} = &D_\theta({\hat o}_{t+k}, a_{t+k}), \forall \, k = 0, \dots, H-1
\end{aligned}
\label{eq:planning_opt}
\end{equation}

In this formulation, $o_{t}$ denotes the current observation, $\hat o_{i}$ is simulated by $D_\theta$, which also incorporates gait commands $g^{cmd}_{t}$, $\gamma$ is the discount factor, and $R_\theta$, $V_\theta$, and $D_\theta$ represent the reward function, value function, and dynamics model, respectively. The objective is to maximize the total expected return, subject to the learned dynamics.

We adopt a MPPI algorithm for deployment due to its sampling-based nature and its ability to operate without requiring gradients. This allows us to leverage the learned dreamer dynamics and reward models even if they are non-smooth or non-differentiable. MPPI also accommodates action and gait constraints directly in the sampling procedure.


At each time step $(t)$, a set of trajectories is sampled from the Dreamer policy and dynamics model, which serves as a warm start to initialize the mean and variance for MPPI. Using this warm-started Gaussian distribution, action sequences are sampled, and corresponding trajectory rollouts are generated through the Dreamer dynamics model. The returns are then computed using the Dreamer reward model with value bootstrapping. The top-performing trajectories are selected based on their returns, and a new Gaussian distribution is fitted to these elites via a momentum-based update. Here, $A_\tau^S$ denotes the sampled action trajectory space, $A_\tau^E$ represents the elite action trajectory space, and $R_\tau$ is the set of returns associated with the action trajectories.

\begin{algorithm}[htp!]
\small
\DontPrintSemicolon
\caption{MPPI Planner with Gait Optimization}
\label{alg:mpc}

$\theta$: Dreamer model parameters; \\
$(\mu_{t-1}, \sigma_{t-1})$: prior distribution parameters; \\
$o_t$: current observation; \\
$g^{cmd}_{\text{prev}}$: previous gait command

\If{$t == 0$}{
    Set $g^{cmd}_{\text{prev}} \leftarrow$ nominal gait (e.g., trot)\;
}

\textbf{Warm start:} Initialize $(\mu_{t-1}, \sigma_{t-1})$, and $M_\pi$ from policy $\pi_\theta$ using dynamics $D_\theta$\;
Set $(\mu^0, \sigma^0) \leftarrow (\mu_{t-1}, \sigma_{t-1})$\;
Initialize empty sets $A^E_\tau, A^S_\tau, R_\tau$\;

\For{$i \leftarrow 1$ \KwTo $N$}{
    Sample $M$ trajectories from $\mathcal{N}(\mu^{i-1}, \sigma^{i-1})$\;

    \For{each sampled trajectory $(g^{j,{cmd}}_t, \dots, g^{j,{cmd}}_{t+H}, a^j_t, \dots, a^j_{t+H})$}{
        \textbf{Wrap:} $(g^{j,{cmd}}_t, \dots, g^{j,{cmd}}_{t+H}) \leftarrow \text{wrap}(g^{j,{cmd}}_t, \dots, g^{j,{cmd}}_{t+H})$\;
        
        $R_j \leftarrow 0$\;

        Set $\hat{o}_t^j \leftarrow$ simulate $o_t$ with $g^{j,{cmd}}_t$\;

        \For{$k \leftarrow 0$ \KwTo $H-1$}{
            $R_j \mathrel{+}= \gamma^k \cdot R_\theta(\hat{o}^j_{t+k}, a^j_{t+k})$\;
            $\hat{o}^j_{t+k+1} \leftarrow D_{\theta}(\hat{o}^j_{t+k}, a^j_{t+k})$\;

            \tcp*[r]{Add gait continuity penalty}
            $R_j \mathrel{-}= \lambda \cdot \| g^{j,{cmd}}_{t+k} - g^{j,{cmd}}_{t+k-1} \|^2$\;
        }

        \tcp*[r]{Add terminal value prediction}
        $R_j \mathrel{+}= \gamma^H V_\theta(\hat{o}^j_{t+H})$\;

        Add $(g^{j,{cmd}}_t, \dots, g^{j,{cmd}}_{t+H}, a^j_t, \dots, a^j_{t+H})$, $R_j$ to $A^S_\tau$, $R_\tau$\;
    }

    Sort $A^S_\tau$ by $R_\tau$ and select top-$M_{\text{elite}}$ trajectories to form $A^E_\tau$\;

    Fit new distribution $(\mu_{\text{elite}}, \sigma_{\text{elite}})$ to $A^E_\tau$\;

    Update $(\mu^i, \sigma^i) \leftarrow \beta(\mu_{\text{elite}}, \sigma_{\text{elite}}) + (1 - \beta)(\mu^{i-1}, \sigma^{i-1})$\;
}

\KwResult{First action from $\mathcal{N}(\mu^N, \sigma^N)$ and updated $g^{cmd}_{\text{prev}}$}
\end{algorithm}

\subsection{Gait Parameter Normalization}

The gait parameters are represented as a $[3 \times 1]$ vector, following the formulation in~\cite{wtw}. Each element is a scalar in the range $[0, 1]$, encoding the phase offset between specific leg pairs. This compact representation enables continuous modulation of gait structures and allows the expression of all two-beat quadrupedal contact patterns. Smooth and bounded gait variation is enforced using a wrap function.

\[
\tilde{g}^{cmd} = (g^{cmd} \space \% \space 1.0)
\]

\subsection{Integrated Architecture}
To summarise, the pipeline operates in two stages:
\begin{itemize}
    \item \textbf{Offline Training:} A dreamer model is trained end-to-end using a diverse, gait-conditioned RL framework. It learns the policy, dynamics, reward, and value functions. The modular structure of dreamer enhances interpretability and flexibility within the framework.
    \item \textbf{Online Deployment:} During deployment, an MPPI controller leverages the learned dreamer model to jointly optimize for both actions and gait parameters.
\end{itemize}

The proposed architecture enables generalization across environments and supports adaptive, energy-efficient locomotion with gait transitions that emerge from optimization rather than manual scripting.

\section{Experiments and Results}

\begin{figure*}[!htp]
    \centering
    \includegraphics[width=0.9\linewidth]{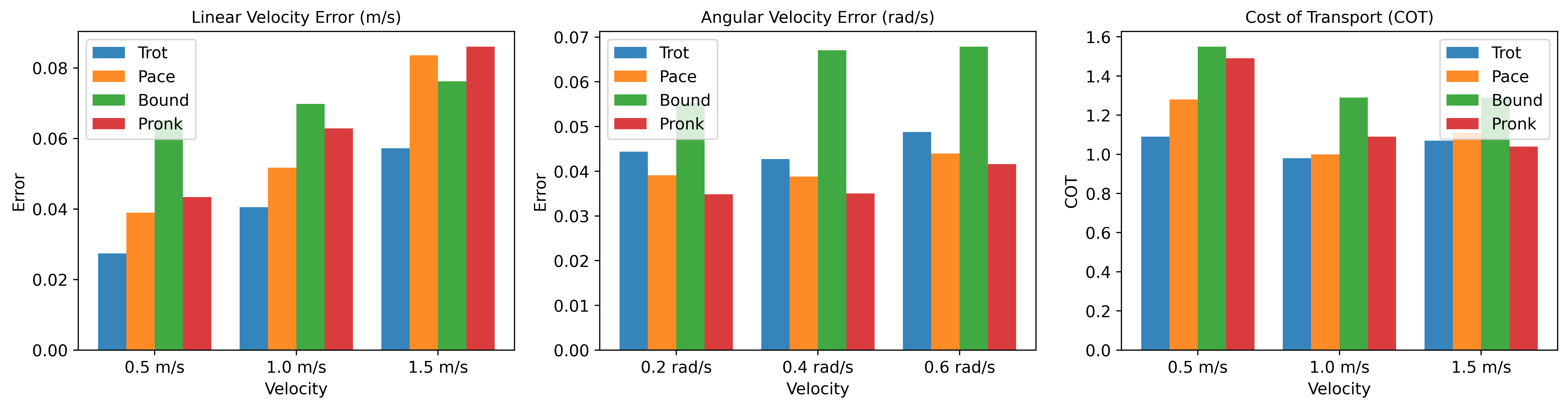}
    \caption{Performance comparison of fixed gaits across command velocities on flat terrain. 
    The results illustrate how linear and angular tracking errors, as well as cost of transport, vary significantly with gait choice—especially at higher velocities—highlighting the limitations of using a single gait policy and the need for adaptive gait selection. \vspace{-10pt}}
    \label{fig:Ablation}
\end{figure*}

We evaluate our framework through a series of experiments in simulation, focusing on algorithm performance, gait transitions, and energy efficiency. The results highlight the limitations of fixed-gait policies and demonstrate the benefits of adaptive gait optimization.

For both training and deployment, we used an open-source simulation environment for the Unitree Go1 quadruped robot, built on NVIDIA’s Isaac Gym simulator~\cite{isaac}. The PPO policy~\cite{ppo} and its corresponding neural network were implemented and trained using PyTorch~\cite{pytorch}, while the dreamer module—including the policy, dynamics model, value function, and reward model, which are Multi-Layer Perceptrons with hidden dimensions (256, 128) was implemented using JAX~\cite{jax}.

\subsection{Ablation study}
To validate the motivation behind our approach, we perform an ablation study that examines the limitations of using fixed gaits in a flat-terrain setting. Specifically, we evaluate four major gaits—trot, pace, bound, and pronk—by measuring velocity tracking error and cost of transport across a range of commanded velocities. This analysis reveals (as shown in fig. \ref{fig:Ablation}) that no single gait performs well across all speeds, highlighting the need for a framework that can optimize gait selection during deployment based on task demands and performance criteria.


From Fig.~\ref{fig:Ablation}, we observe that different gaits exhibit distinct performance profiles as target speed increases. Trotting maintains consistently low linear velocity errors; however, as the speed rises from 0.5 m/s to 1.5 m/s, its Cost of Transport (CoT) remains nearly constant. In contrast, pronking becomes more energy-efficient at higher speeds while maintaining competitive tracking accuracy, suggesting better suitability for fast locomotion. These observations highlight that no single gait excels across all conditions, underscoring the need for a planning framework that adaptively selects gaits based on task demands and performance trade-offs.

\begin{figure*}[!htp]
    \centering
    \includegraphics[width=.7\linewidth]{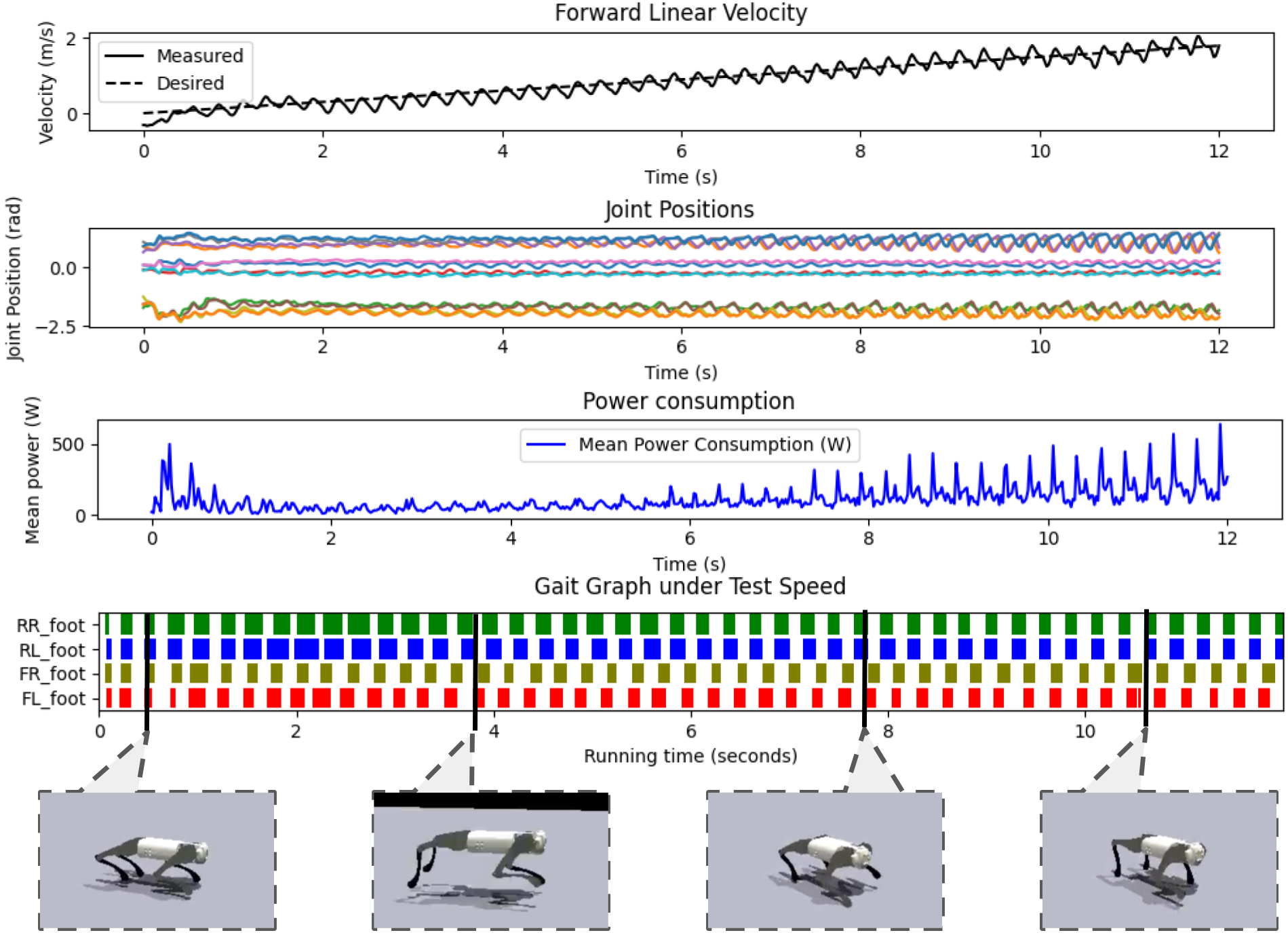}
    \caption{Evaluation of the proposed locomotion planner under a continuously increasing velocity command. The robot effectively tracks the desired forward velocity with smooth acceleration. Joint trajectories exhibit stable and periodic behavior, indicating coordinated motion. Mean power consumption shows an initial peak during acceleration followed by steady oscillations. The gait diagram illustrates consistent foot contact patterns and adaptive gait transitions corresponding to the increasing command speed, with representative snapshots shown below. \vspace{-10pt}}
    \label{fig:gait-change}
\end{figure*}

\begin{table}[htp!]
\scriptsize
\centering
\caption{Tunable hyperparameters of Algorithm~\ref{alg:mpc}.}
\label{tab:mpc}
\setlength{\tabcolsep}{5pt}
\renewcommand{\arraystretch}{1.3}

\begin{minipage}[t]{0.48\linewidth}
\centering
\begin{tabular}{l|c}
\toprule
\textbf{Hyperparameter} & \textbf{Value} \\
\midrule
Horizon ($H$) & 6 \\
MPC iterations ($N$) & 6 \\
Num. samples ($M$) & 500 \\
Num. policy samples ($M_\pi$) & 30 \\
\bottomrule
\end{tabular}
\end{minipage}
\hfill
\begin{minipage}[t]{0.48\linewidth}
\centering
\begin{tabular}{l|c}
\toprule
\textbf{Hyperparameter} & \textbf{Value} \\
\midrule
Num. elite samples ($M_{\text{elite}}$) & 60 \\
Discount factor ($\gamma$) & 0.99 \\
Momentum factor ($\beta$) & 0.95 \\
Temperature ($T$) & 0.5 \\
\bottomrule
\end{tabular}
\end{minipage}
\vspace{-15pt}
\end{table}

\subsection{Gait Adaptation}



The simulation results are shown in Fig.~\ref{fig:gait-change}, where a continuously increasing forward velocity command is used to evaluate how the planner adapts its gait in response to gradually changing task demands. The robot successfully tracks the commanded velocity throughout the episode, and the joint position trajectories remain cyclic and smooth, indicating stable locomotion even as the gait evolves over time.

To visualize gait modulation, we directly plot the foot contact patterns for all four legs. These contact schedules reveal how interlimb coordination adapts with increasing velocity, capturing continuous gait transitions without needing discrete gait labels, also highlighting the planner's ability to generate context-specific gaits in a continuous manner.

A key observation is that the robot maintains reliable velocity tracking performance during these transitions, with no visible degradation in motion stability or quality. The smooth joint trajectories further confirm that transitions between gaits are not abrupt or disruptive. Although trotting is commonly favored in RL-based quadrupedal locomotion, the planner frequently adapts to alternative coordination patterns depending on the context—suggesting that tracking accuracy and energy efficiency are better achieved through continuous and adaptive gait modulation. These results demonstrate that the planner effectively selects task-appropriate gaits while preserving motion continuity and robustness.

\begin{table}[htp!]
\vspace{10pt}
\centering
\caption{Cost of transport comparison}
\label{tab:energy_results}
\scriptsize
\setlength{\tabcolsep}{5pt}
\renewcommand{\arraystretch}{1.3}
\begin{tabular}{l|c|c|c|c}
\toprule
\textbf{Methods / Velocity} & 0.5 m/s & 1.0 m/s & 1.5 m/s & 2.0 m/s \\
\midrule
\textbf{RL: Trotting} & 1.09 & 0.98 & 1.07 & 1.16  \\
\textbf{RL: Pacing} & 1.28 & 1.00 & 1.11 & 1.19  \\
\textbf{RL: Bounding} & 1.55 & 1.29 & 1.29 & 1.28  \\
\textbf{RL: Pronking} & 1.49 & 1.09 & 1.04 & 1.08 \\
\textbf{Ours} & \textbf{0.918} & \textbf{0.771} & \textbf{0.822} & \textbf{0.914} \\
\bottomrule
\end{tabular}
\end{table}


\subsection{Energy Consumption}

We analyzed energy consumption by comparing our gait adaptation method with fixed-gait RL baselines using Cost of Transport (CoT), defined as $ \frac{|P|}{m \cdot g \cdot v}$. Here, $|P|$ denotes the average power (time-averaged sum of joint torque–velocity products across all joints), $m$ is the robot mass, $g$ is gravitational acceleration, and $v$     is the average forward velocity.


As shown in Table~\ref{tab:energy_results}, our method consistently results in a lower cost of transport across all commanded velocities. Notably, the energy consumption remains 15-20\% lower constantly when compared to the most efficient gaits in each velocity range.

This result indicates the inefficiency of using a single gait across all speeds. For example, trotting is efficient at lower speeds, whereas pronking performs better at 2.0 m/s. Our framework avoids such mismatches by dynamically adapting the gait to suit the environment, resulting in smoother transitions and significantly reduced energy expenditure.
\vspace{-5pt}
\subsection{Discussion and Future Work}

Some of the practical considerations for hardware deployment and feasibility of the proposed method include computational overhead and onboard inference frequency. Although the results presented above were obtained using a high-specification server, our framework can be deployed on relatively smaller systems as well. Specifically, we used an NVIDIA A6000 GPU, which required approximately 5–6 hours to train the RL policy with the Dreamer module. For inference, we evaluated the controller on an independent system equipped with an NVIDIA RTX 3080 GPU, achieving a control frequency of around 330~Hz. This indicates the potential for smooth hardware deployment even on moderately powered systems. Onboard GPU computation remains a requirement for real-world deployment.


The prospect of hardware deployment motivates extending this framework to multi-terrain locomotion, where quadrupeds must continuously adapt their gait to uneven and rough surfaces for robust performance. Future work will focus on developing terrain-aware extensions of the proposed method, incorporating visual inputs for predictive planning, and integrating the Dreamer module with structured models such as Lagrangian Neural Networks (LNNs) to enhance generalization and reliability in real-world deployment.

\vspace{-5pt}
\section{Conclusion}

In this work, we proposed an optimization framework for adaptive quadrupedal locomotion that selects gaits based on task demands such as speed and energy efficiency, rather than relying on a fixed policy. Unlike standard RL approaches that often converge to a single gait, our method optimizes gait parameters online using Model Predictive Path Integral (MPPI) control. Simulation results show that our approach maintains high velocity tracking accuracy across a wide range of commanded speeds while reducing energy consumption by up to 40\% at higher speeds compared to fixed-gait baselines. The results further demonstrate smooth transitions between gaits without compromising performance, highlighting the framework’s effectiveness in achieving both adaptability and efficiency.

\bibliographystyle{ieeetr}
\bibliography{references}

\end{document}